\documentclass[fleqn,10pt]{wlscirep}
\usepackage[utf8]{inputenc}
\usepackage[T1]{fontenc}
\usepackage{bm}
\usepackage{xcolor}
\usepackage{graphicx}
\usepackage{amsmath}
\usepackage{tikz}
\usetikzlibrary{shapes.geometric, arrows.meta, calc, fit, positioning, decorations.markings, decorations.pathmorphing, backgrounds} 

\newcommand{\Heven}{H_{\mathrm{even}}}
\newcommand{\Hodd}{H_{\mathrm{odd}}}
\newcommand{\PhiTopo}{\Phi_{\mathrm{Topo}}}
\newcommand{\CH}{C_H}
\newcommand{\bbeta}{\beta}
\newcommand{\EulerChi}{\chi}

\title{On the Topological Foundation of Learning and Memory}

\author[1,*]{Xin Li}
\affil[1]{Department of Computer Science, University at Albany, Albany, NY 12222}

\affil[*]{e-mail: xli48@albany.edu}

\begin{abstract}
We propose a formal foundation for cognition rooted in algebraic topology, built on a Homological Parity Principle. This posits that even-dimensional homology ($\Heven$) represents stable Structure/Context (e.g., generative models), while odd-dimensional homology ($\Hodd$) represents dynamic Flow/Content (e.g., sensory/memory data). Cognition is governed by the Context-Content Uncertainty Principle (CCUP), a dynamical cycle aligning these parities. This framework distinguishes two modes: Inference (waking), where the Heven scaffold predicts the Hodd flow (a Context-before-Content process); and Learning (sleep), an inverted Structure-before-Specificity process where Hodd memory traces sculpt the Heven scaffold. This parity interpretation unifies cognitive functions like semantic ($\Heven$) and episodic ($\Hodd$) memory and provides a structural generalization of existing theories, recasting Friston's Free Energy Principle and Tonini's Integrated Information in topological terms.
\end{abstract}

\begin{document}

\flushbottom
\maketitle


Grand unified theories of cognition, such as the Free Energy Principle (FEP) \cite{friston2010free} or Integrated Information Theory (IIT) \cite{tononi2016integrated}, seek fundamental principles of perception and consciousness. While FEP describes the \textit{dynamics} of inference (minimizing surprisal) and IIT quantifies the \textit{capacity} for consciousness (a scalar, $\Phi$), neither fully describes the underlying \textit{structure} of the representational space itself. We propose this missing foundation is topological, and that its organization follows a fundamental parity-based principle derived from the Euler characteristic, $\chi$. 
The foundational axiom of algebraic topology, $\partial^2=0$ (the boundary of a boundary is nil) \cite{wheeler1990information}, ensures that homology groups ($\beta_k = \dim H_k$) are well-defined and gives rise to the Euler-Poincaré formula, $\chi = \sum_k (-1)^k \beta_k$ \cite{hatcher2005algebraic, nakahara2018geometry}. We propose a physical interpretation of this invariant based on its decomposition by parity: $\chi = (\sum \beta_{2k}) - (\sum \beta_{2k+1})$, or $\chi = \Heven - \Hodd$. This $\Heven - \Hodd$ formulation provides the central motivation for our homological framework. 

We identify the even-dimensional $\Heven$ with the stable, scaffold of a cognitive system, representing the generative \textbf{world model} \cite{brown2020language, ha2018world} and its associated semantic knowledge \cite{binder2011neurobiology}. Conversely, we identify the odd-dimensional $\Hodd$ with the dynamic, flow of internal processes, representing the \textbf{latent self} (e.g., episodic traces \cite{tulving1972episodic}, active inference \cite{friston2017active}). The Euler characteristic, $\chi = \Heven - \Hodd$, is recast as a measure of the topological \textit{imbalance} between the agent's world-model $\Heven$ (context/scaffold) and its internal, self-representational flow $\Hodd$ (content). The convergence of the context-content alignment cycle \textit{is} the process of minimizing this imbalance. This reformulation provides a topological basis for FEP \cite{friston2010free}, suggesting a self-organizing system maintains its integrity by driving the system towards a state of global coherence where $\chi \approx 0$, which we define as being topologically self-consistent.

We posit that cognition is a dynamical cycle, obeying the Context-Content Uncertainty Principle (CCUP) \cite{robertson1929uncertainty}, that seeks to align these parities through two primary modes (Fig. \ref{fig:principle}) \cite{buzsaki2006rhythms}. The first is \textit{inference} (waking), a context-before-content cycle where the $\Heven$ scaffold provides top-down predictions to constrain the $\Hodd$ flow. This directly formalizes hierarchical Bayesian inference \cite{rao1999predictive}, with perception as the convergence of the cycle, where the $\Hodd$ flow ``snaps'' to the nearest low-energy state on the $\Heven$ scaffold. The second mode is \textit{learning} (sleep) \cite{buzsaki1996hippocampo}, an inverted Structure-before-Specificity (SbS) process. With external sensory noise absent, $\Hodd$ content, dominated by replayed episodic memory traces \cite{wilson1994reactivation}, provides training data to anneal the $\Heven$ scaffold. 
We can abstract the consolidation mechanism \cite{squire2015memory} of the learning mode by the following homological memory model. Let $\gamma_i$ denote the $i$-th memory trace (an $H_{odd}$ flow); it is decomposed as:
$\gamma_i = \sigma + \sum_k a_{ik}\,\beta_k + \partial d_i$,
where $\sigma \in Z_k$ is the context backbone ($\Heven$ scaffold) common to multiple traces; $\beta_k$ are the independent recurrent content loops ($\Hodd$ episodic flow); and $\partial d_i$ is the residual boundary representing transient, unbound noise or surprisal. The SbS learning process is a boundary evolution that minimizes this noise ($\partial d_i \to 0$), leaving the stable memory manifold $\sigma + \sum_k a_{ik}\beta_k$ and thereby refining the $\Heven$ scaffold ($\sigma$).

The proposed homological memory model elegantly explains several core dissociations in cognition. The distinction between stable semantic memory ($\Heven$) and dynamic episodic memory ($\Hodd$) \cite{tulving1972episodic} becomes a foundational feature of the architecture. Likewise, it maps the rules of syntax ($\Heven$ scaffold) to the flow of meaning ($\Hodd$ process), providing a topological basis for semantic bootstrapping \cite{pinker2014bootstrapping}. In social cognition, it separates the $\Heven$ scaffold (Cognitive Theory of Mind) from the $\Hodd$ flow (Affective Theory of Mind) \cite{shamay2009two}.

Our topological framework also provides a structural generalization of existing theories \cite{tononi2016integrated}. It recasts consciousness not as a scalar ($\Phi$) but as a topological signature of parity balance. A system's integrated information can be reformulated as $\PhiTopo = \CH - |\EulerChi|$, where $\CH = \sum_k \bbeta_k$ is the total topological richness and $\EulerChi$ is the Euler characteristic. This value is maximized only when the system has both rich homology (high $\CH$) and global coherence (low $|\EulerChi|$) \cite{ayzenberg2025sheaf}, meaning its $\Heven$ scaffold and $\Hodd$ flow are equally balanced. This model also complements FEP \cite{friston2010free} by defining the very \textit{landscape} upon which active inference unfolds: FEP describes the $H_{odd}$ flow of beliefs descending a gradient, while homology describes the $H_{even}$ structure of the generative model that creates the gradient. In short, this topological parity principle offers a unifying foundation, linking the geometry of mental models to the dynamics of learning, memory, and perception \cite{izhikevich2006polychronization}.

We propose a plausible neurological substrate for the advocated parity principle. The stable $\Heven$ scaffold corresponds to the brain's anatomical and slow-changing functional connectome, the synaptic-weighted architecture of cortical columns that encode the generative model \cite{petri2014homological}. The dynamic $\Hodd$ flow, in contrast, corresponds to the transient, high-frequency process of neural activity, such as the phase-locked cycles of theta-gamma oscillations \cite{canolty2006high} that run on that scaffold \cite{buzsaki2006rhythms, izhikevich2006polychronization}. This maps directly to the homological memory model: a ``flash of insight'' or ``Aha!'' moment is the emergence of a new, persistent $H_{odd}$ oscillatory pattern ($\sigma_i$) that is not yet supported by the underlying $\Heven$ scaffold. The subsequent SbS learning process (i.e., sleep \cite{buzsaki1996hippocampo}) is the synaptic plasticity (LTP) that physically hardens this new $\Hodd$ cycle into the $\Heven$ connectome, updating the world model by making the novel insightful ($\sigma_i$) part of the stable known basis ($\beta_j$).

Finally, the proposed scaffold-flow model generates specific, falsifiable predictions for both healthy and pathological cognition. We predict that a state of high free energy (i.e., high surprisal) will correspond to a high degree of topological imbalance ($|\chi| \gg 0$). This suggests a novel class of topological biomarkers for psychiatric disorders. For example, disorders of psychosis (e.g., schizophrenia), often framed as a failure of priors where the generative model fails to constrain sensory flow \cite{friston1998disconnection}, would be characterized by a chaotic state where $\Hodd \gg \Heven$, resulting in a large negative Euler characteristic. Conversely, disorders of rigidity (e.g., Autism Spectrum Disorder \cite{wang2023revisit}), often framed as overly-precise priors that are resistant to sensory updating \cite{van2014precise}, would be characterized by a rigid state where $\Heven \gg \Hodd$, resulting in a large positive Euler characteristic. This framework also suggests new avenues for AI, framing catastrophic forgetting in artificial neural networks (ANNs) as a failure to build a stable $\Heven$ scaffold, and proposing that a sleep-wake cycle \cite{fuller2006neurobiology} that uses the homological memory model for offline consolidation could resolve it. We conclude that this dual-mode architecture, which uses a sleep phase ($\Hodd$ flow) to anneal its waking scaffold ($\Heven$), is not a quirk of biological cognition but a necessary topological foundation for any system, artificial or natural, to achieve generalizable intelligence \cite{bubeck2023sparks}.

\begin{figure}[t]
\centering
\fbox{\begin{minipage}{0.9\textwidth}
\textbf{The Homological Parity Principle}

\noindent The principle posits a fundamental dichotomy in cognitive topology. \textbf{Even-dimensional homology ($\Heven$)}, such as components ($\bbeta_0$) and cavities ($\bbeta_2$), represents the stable, time-invariant \textbf{scaffold, structure, or context} of a system (e.g., a generative model, semantic rules). Conversely, \textbf{Odd-dimensional homology ($\Hodd$)}, such as cycles ($\bbeta_1$) and voids ($\bbeta_3$), represents the dynamic, time-varying \textbf{flow, process, or content} that unfolds upon that structure (e.g., a sensory stream, an episodic memory, a conscious thought).
\end{minipage}}
\begin{tikzpicture}[
    >=Latex,
    font=\small,
    backbone/.style={line width=0.9pt},
    loop/.style={line width=0.8pt,dashed},
    flow/.style={-Latex, line width=0.8pt},
    anchorpt/.style={circle, fill=black, inner sep=0pt, minimum size=2.2pt},
    tag/.style={fill=black!4, draw, rounded corners, inner sep=2pt},
    lab/.style={font=\scriptsize, align=center}
]

\def\xL{0.5}
\def\xR{12.5}
\coordinate (S0) at (\xL,0);
\coordinate (S4) at (\xR,0);

\coordinate (S1) at ($(S0)!0.30!(S4)$);
\coordinate (S2) at ($(S0)!0.55!(S4)$);
\coordinate (S3) at ($(S0)!0.80!(S4)$);

\draw[backbone] (S0)--(S4);
\node[lab,above] at ($(S0)!0.5!(S4)+(3.5,-0.5)$)
  {$\sigma$ (even-dimensional scaffold: shared, reusable backbone)};

\node[lab,anchor=south west] at ($(S0)+(-0.15,0.20)$) {(a) scaffolds};

\node[anchorpt] (A1) at (S1) {};
\node[anchorpt] (A2) at (S2) {};
\node[anchorpt] (A3) at (S3) {};

\path let \p1=(A1) in coordinate (B1L) at (\x1-0.9,1.05) coordinate (B1R) at (\x1+0.9,1.05);
\draw[loop] (A1) .. controls ($(A1)+(0,0.55)$) and (B1L) .. ($(A1)+(0,1.05)$)
            .. controls (B1R) and ($(A1)+(0,0.55)$) .. (A1);
\node[lab,above] at ($(A1)+(0,1.25)$) {$b_{1}$};

\path let \p2=(A2) in coordinate (B2L) at (\x2-1.15,1.35) coordinate (B2R) at (\x2+1.15,1.35);
\draw[loop] (A2) .. controls ($(A2)+(0,0.70)$) and (B2L) .. ($(A2)+(0,1.35)$)
            .. controls (B2R) and ($(A2)+(0,0.70)$) .. (A2);
\node[lab,above] at ($(A2)+(0,1.60)$) {$b_{2}$};

\path let \p3=(A3) in coordinate (B3L) at (\x3-1.00,1.10) coordinate (B3R) at (\x3+1.00,1.10);
\draw[loop] (A3) .. controls ($(A3)+(0,0.60)$) and (B3L) .. ($(A3)+(0,1.10)$)
            .. controls (B3R) and ($(A3)+(0,0.60)$) .. (A3);
\node[lab,above] at ($(A3)+(0,1.30)$) {$b_{3}$};

\node[lab,anchor=south] at ($(A2)+(0,2.0)$) {(b) flows};

\draw[flow] ($(S0)!0.42!(S2)$) -- node[lab,below,xshift=-50pt,yshift=-1pt]
  {$\mathcal{R}$: re-enter scaffold $\sigma$} ($(S0)!0.58!(S2)$);

\draw[flow,->] ($(A2)+(-0.25,-0.08)$)
  to[bend left=18] ($(A2)+(0.05,0.80)$);
\node[lab,left] at ($(A2)+(-0.35,0.45)$) {$\mathcal{F}$};

\draw[flow,->] ($(A2)+(0.05,0.80)$)
  to[bend left=40] ($(A2)+(0.25,0.00)$);
\node[lab,right] at ($(A2)+(0.35,0.45)$) {$\mathcal{R}$};

\draw[flow] ($(A1)+(0.00,0.95)$) arc[start angle=90,end angle=-230,radius=0.95cm];

\node[tag, below=10mm of $(S0)!0.5!(S4)$] (leg) {%
\begin{minipage}{0.94\linewidth}
\footnotesize
\textbf{Legend.} \textbf{(a) Even-dimensional homology as scaffolds.}
$\sigma$ denotes a shared, low-entropy backbone (e.g. classes in $H_{0},H_{2}$)
that binds and integrates content across contexts. 
\textbf{(b) Odd-dimensional homology as flows.}
$b_k$ denote admissible recurrent loops (e.g. classes in $H_{1},H_{3}$) that ride
on $\sigma$ and encode invariants over time.
Homological parity separates \emph{structure} (even, scaffolds) from
\emph{dynamics} (odd, flows), while $\partial^{2}=0$ enforces re-closure of
flows onto their supporting scaffolds.
\end{minipage}
};

\node[lab,anchor=west] at ($(S4)+(-1.2,0.95)$)
  {even: scaffolds,\; odd: flows,\; $\partial^{2}=0$};

\end{tikzpicture}

\caption{The core principle mapping cognitive function to topological parity and the scaffold-flow model constructed based on the parity principle.}
\label{fig:principle}
\end{figure}

\vspace{0.5cm}
\noindent\textbf{Acknowledgements}\\
This work was partially supported by NSF IIS-2401748 and BCS-2401398.\\

\noindent\textbf{Competing interests}\\
The authors declare that they have no known competing financial interests or personal relationships that could have appeared to influence the work reported in this paper. \\

\bibliography{ref}

\end{document}